\documentclass{article}

\usepackage{iclr2026_conference,times}
\usepackage{amsmath,amssymb,amsthm}
\usepackage{booktabs}
\usepackage{array}
\usepackage{graphicx}
\usepackage{hyperref}
\usepackage{url}
\usepackage{enumitem}

\iclrfinalcopy

\newtheorem{lemma}{Lemma}

\newcommand{\E}{\mathbb{E}}
\newcommand{\R}{\mathcal{R}}
\newcommand{\F}{\mathcal{F}}
\newcommand{\J}{\mathcal{J}}
\newcommand{\Profile}{\operatorname{profile}}

\title{Stopping and Routing LLM Judge Panels}
\author{
Bin Zhu \quad Yi Xie \quad Yanghui Rao\thanks{Corresponding author.}\\
School of Computer Science and Engineering\\
Sun Yat-sen University, Guangzhou, China\\
\texttt{zhub35@mail2.sysu.edu.cn \quad xiey299@mail2.sysu.edu.cn}\\
\texttt{raoyangh@mail.sysu.edu.cn}
}
\date{}

\begin{document}
\maketitle

\begin{abstract}
LLM evaluation pipelines often have many candidate judges: general
LLM-as-a-judge prompts, reward models, safety classifiers, confidence variants,
and task-specific verifiers. The deployment question is not only which judge is
best, but which judges should be called, on which examples, and when panel
construction should stop. We formulate judge-panel design as a
role-conditioned allocation problem. From a small labeled audit set, declared
slices, and judge costs, the method estimates target-relative roles:
\emph{copies} add no conditional information, \emph{complements} improve the
global panel, and \emph{specialists} help only on slices. These roles induce a
policy: drop copies, add complements globally, route specialists conditionally,
and stop when validation gain falls below a threshold. Across reasoning, code,
safety, preference, reward-model, summarization, and math audits, the method is
compared with single judges, flat panels, matched diversity heuristics,
full-call stacking, reliability juries, and frugal cascades. The result is a
regime map for judge calls: route specialists on deployable slices, stop in
saturated verifier regimes, keep broad ensembles when their risk benefit is
worth the cost, and ignore conditional copies. The output is a reusable,
auditable call plan for the next evaluation batch.
\end{abstract}

\section{Introduction}

LLM-as-a-judge systems are now common in model evaluation
\citep{zheng2023judging,liu2023geval,zhu2023judgelm,kim2024prometheus2,
verga2024juries,kocmi2023state,dubois2024length,chaneval2024}. A realistic
evaluation pipeline may include a general judge, a rubric prompt, a reward
model, a safety classifier, a confidence variant, and a deterministic verifier.
For every new evaluation batch, the researcher must make a concrete operating
decision: call the whole panel, call a cheap verifier and stop, route a safety
judge only to risky cases, or drop a redundant prompt entirely. A static judge
ranking does not answer that question. The value of a judge is conditional on
the current panel, the target distribution, and the slice of examples where it
will be used.

This creates a useful opportunity. A small labeled audit set can turn judge
diversity from a descriptive property into a calling policy. A safety judge can
become a specialist on jailbreak failures; a verifier can make several LLM
judges redundant; and a full ensemble can still be the right endpoint on broad
math or reward-model regimes \citep{wolpert1992stacked,caruana2004ensemble}.
The goal is to identify these cases before paying for judge calls on the next
batch.

We turn the taxonomy of \emph{copy}, \emph{complement}, and \emph{specialist}
into an allocation method. The output is a calling policy
\(\pi(x)\subseteq\J\) with a validation-based stopping record.
The method asks whether each candidate reduces held-out calibration risk after
conditioning on the current panel, and whether that gain is global or
slice-specific. Copies are dropped, broad complements are added to the global
panel, specialists are routed to their slices, and construction stops when no
remaining candidate clears a declared gain threshold.

\paragraph{Contributions.}
We make three claims. First, judge diversity should be target-relative and
conditional, not nominal. Second, copy/complement/specialist roles can be
converted into a practical policy with costs, slices, and stopping conditions.
Third, the empirical value of the method is a regime map for deployment: it
identifies when to route specialists, when to keep a cheap stopped panel, when
to stop after a verifier, and when to pay for the full panel.

Although this paper and the companion \emph{A Finite-Calibration Regime Map for
LLM Judge Panels} share part of the benchmark judge-output matrices and judge
pool, they address distinct deployment decisions: this paper selects
conditional calls and stopping, whereas the companion selects a panel prefix
and aggregation family after candidate outputs are available.

\section{Role-Conditioned Allocation}

Let \(X\) be an evaluated item and \(Y\in[0,1]\) the audit label. A finite
candidate pool \(\J\) contains judge signals \(Z_j\) such as
\texttt{correct}/\texttt{incorrect}, \texttt{safe}/\texttt{unsafe}, or
\texttt{A}/\texttt{B}. For a panel \(S\subseteq\J\), let \(Z_S\) be the joint
output pattern. The researcher declares slices \(\F\) that matter for the
target distribution, such as LLMBar subsets \citep{zeng2024llmbar}, safety
failure modes \citep{chao2024jailbreakbench}, generator type, or difficulty
level \citep{hendrycks2021math}. The goal is a policy
\(\pi(x;\J,\F)\subseteq\J\) that decides which judges to call on \(x\).

Fix a target distribution \(P\). For a panel \(S\), define the oracle predictor
\[
\eta_{P,S}(z)=\E_P[Y\mid Z_S=z]
\]
and its squared-loss oracle risk
\[
\R^\star_{P,S}
=
\E_P[(Y-\eta_{P,S}(Z_S))^2].
\]
The conditional value of adding judge \(j\notin S\) is
\[
g_P(j\mid S)
=
\R^\star_{P,S}-\R^\star_{P,S\cup\{j\}}.
\]
This is the target information in \(j\) that is not already present in \(S\).

\begin{lemma}[Projection gain identity]
For any finite panel \(S\) and candidate judge \(j\notin S\),
\[
g_P(j\mid S)
=
\E_P\left[
\left(
\eta_{P,S\cup\{j\}}(Z_{S\cup\{j\}})
-
\eta_{P,S}(Z_S)
\right)^2
\right]\ge 0.
\]
The identity requires no independence assumption among judges.
\end{lemma}

For slices \(f\in\F\), define broad gain \(C_P(j\mid S)=g_P(j\mid S)\) and
slice gain \(A_f(j\mid S)=g_{P_f}(j\mid S)\). The role profile
\(\Profile_{P,\F,S}(j)=(C_P,\{A_f\}_{f\in\F})\) is multi-label: a judge may be
both a broad complement and a slice specialist, and its role can change after
another judge enters the panel.

The profile is deliberately an action interface rather than a naming scheme.
For example, a reward model that is weak as a standalone preference judge can
still be a complement after a rubric prompt enters the panel if it separates
cases the prompt collapses. Conversely, a second prompt from the same model
family can become a copy if its conditional gain vanishes after the first
prompt. Slice roles are evaluated in the same target-relative way. We also
track a diagnostic specialization ratio
\[
\rho_f(j\mid S)
=
\frac{g_{P_f}(j\mid S)}{g_P(j\mid S)+\epsilon_0},
\]
with \(\epsilon_0>0\) only to avoid division by zero. The ratio does not make
roles mutually exclusive; it flags concentration of value. A judge may be a
broad complement and still be especially worth inspecting on one declared
slice.

\begin{figure}[t]
\centering
\includegraphics[width=0.78\linewidth]{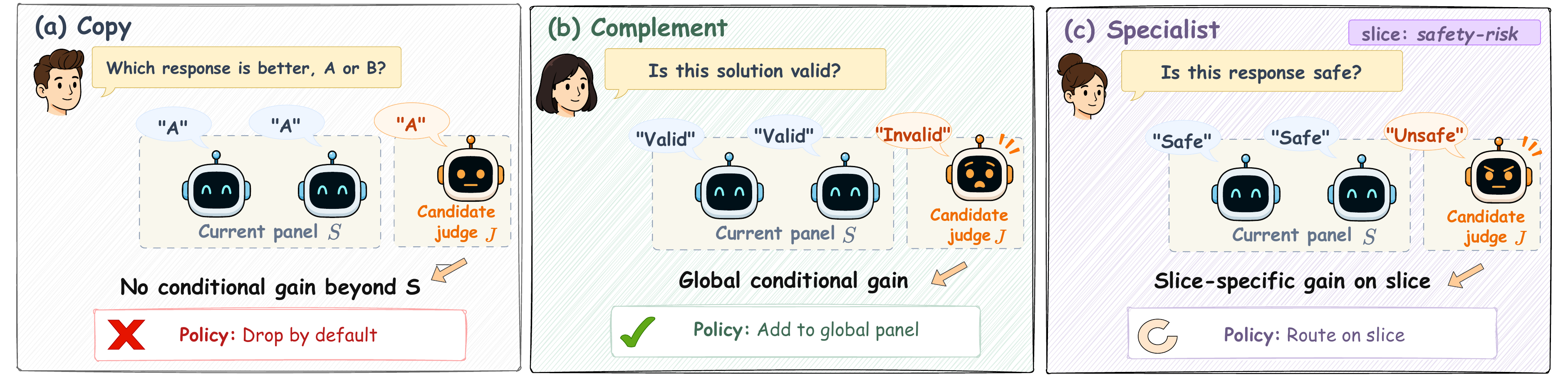}
\caption{Action-oriented role taxonomy. A copy is redundant after conditioning
on the current panel, a complement adds broad residual information, and a
specialist adds value mainly on a deployable slice.}
\label{fig:role-taxonomy}
\end{figure}

\begin{table}[t]
\centering
\small
\begin{tabular}{p{0.16\linewidth}p{0.36\linewidth}p{0.34\linewidth}}
\toprule
Role & Signal pattern & Policy implication \\
\midrule
Copy &
Broad and slice gains are below threshold. &
Do not invoke by default. \\
Complement &
Broad gain \(C_P(j\mid S)\) is above threshold. &
Add to the global panel. \\
Specialist &
Cost-adjusted slice gain clears the slice threshold. &
Route to examples in the corresponding slice. \\
Comp. + spec. &
Broad gain is high and concentrated on one or more slices. &
Invoke globally; optionally prioritize on the specialist slice. \\
\bottomrule
\end{tabular}
\caption{Role taxonomy as a policy interface. Roles are target-relative,
conditional on the current panel, and may overlap.}
\label{tab:roles}
\end{table}

\paragraph{Construction rule.}
We split each audit set into construction-fit, construction-validation, and
final-test parts. Pattern calibrators estimate \(\eta_{P,S}\) by cell means on
canonicalized joint judge-output patterns on the fit split; unseen validation
or test patterns fall back to the fit-split label mean. Selection uses
validation gain only; reported results use the final-test split only. Given
current global panel \(S\), costs \(c_j\), and
threshold \(\tau_P\), add a global judge only if
\[
\max_{j\in J\setminus S}
\left[
\widehat g^{\mathrm{val}}_P(j\mid S)-\lambda c_j
\right]
>
\tau_P.
\]
For each slice, add a routed specialist only if
\[
\max_{j\in J\setminus (S\cup S_f)}
\left[
\widehat g^{\mathrm{val}}_{P_f}(j\mid S\cup S_f)-\lambda_f c_j
\right]
>
\tau_f.
\]
The deployed policy invokes
\(\pi(x)=S\cup S_{f(x)}\) for examples in slice \(f(x)\), and \(S\) otherwise.
The slice function used in deployment must be computable before the routed
judge call. Ground-truth labels may define audit strata for analysis, but they
are not valid inputs to \(\pi(x)\) on a new example; deployable routes must use
metadata, verifier outputs, classifier outputs, or already-observed judge
disagreement.
If no remaining candidate clears threshold, the policy stops and records a
validation-based stopping report that every unused broad or slice gain is below
the declared threshold. This report is operational rather than asymptotic: it says that,
under the audit split and cost model, the panel is usable without further judge
calls.

Algorithmically, global construction is a greedy validation procedure. Starting
from an empty or user-seeded panel, we fit the current pattern calibrator, score
each unused candidate by cost-adjusted validation gain, add the best candidate
only if it clears \(\tau_P\), and repeat. After the global panel stops, each
slice runs the same greedy search with the selected global panel fixed. The
final policy is then refit on the full construction split and evaluated once on
held-out final-test examples. The stopping report is the collection of
failed inequalities for unused broad and routed candidates. It records the
decision actually made by the deployment policy: under the finite audit split,
declared slices, thresholds, and costs, no remaining single judge call is worth
adding to the current plan. If a deployment owner wants to search for pairwise
or higher-order complementarity, the same validation-gain objective can be run
with beam or subset proposals; the stopping report then documents that expanded
search space.

This finite-split design makes the policy usable as a deployment audit. Judge
selection happens on construction data, final-test examples are held out for
reporting, and the comparisons include both cheap baselines and full-call
aggregation endpoints. The intended use is simple: before paying for future
judge calls, use a labeled audit set to decide whether a candidate adds
information conditional on the panel that will actually be invoked.

\section{Experimental Protocol}

We evaluate non-saturated settings where a single judge is not already perfect:
hard GSM8K rationale audits \citep{cobbe2021training}, MBPP public-test
overfit audits \citep{austin2021program}, JailbreakBench safety
\citep{chao2024jailbreakbench}, LLMBar preference under DeepSeek, Qwen3, and
JudgeLM anchors \citep{zeng2024llmbar,deepseekai2024v3,qwen2025qwen3,
zhu2023judgelm}, RewardBench
\citep{lambert2024rewardbench}, Arena100K \citep{chiang2024chatbot}, SummEval
\citep{fabbri2021summeval}, and MATH-500
\citep{hendrycks2021math,lightman2023lets}. HumanEval and ordinary GSM8K are
used only as saturated stopping checks
\citep{chen2021evaluating,cobbe2021training}. The concrete pool uses Qwen2.5
Instruct 7B, Llama 3.1 Instruct 8B, Mistral v0.3 7B, Prometheus 2 v2.0 7B,
Gemma 3 IT 12B, Atla Selene Mini (Llama 3.1, 8B), and the DeepSeek V4 Flash
API model (284B total parameters, 13B active parameters), with task-specific
subsets where noted. LLM judge calls have normalized cost \(1.0\), and
deterministic verifiers have cost \(0.1\). Route keys are treated as
pre-available metadata, verifier outputs, classifier outputs, or already-observed
judge signals; an additional model call needed to obtain a route key must be
added to the cost model.

Baselines cover the options a practitioner would plausibly deploy: single best
validation judge; flat all-judge panels; matched-\(K\) top-\(k\), correlation
diversity, and quality-diversity panels; full-call ridge/logistic stacking
\citep{wolpert1992stacked,caruana2004ensemble}; Dawid--Skene-style reliability
juries \citep{dawid1979maximum,bradley1952rank,raykar2010learning,
whitehill2009whose}; and FrugalGPT/RouteLLM-style confidence cascades
\citep{chen2024frugalgpt,ong2025routellm}. All results are averaged over 10
random splits; Appendix~\ref{app:split-variation} reports split-level standard
deviations and 95\% confidence intervals for the main risk comparisons. Unless
noted, \(\tau_P=\tau_f=0.005\). Appendix~\ref{app:judge-pools} lists the
judge pools, route keys, and cheap verifiers used in each setting.

\begin{table}[t]
\centering
\scriptsize
\setlength{\tabcolsep}{3.6pt}
\begin{tabular}{p{0.17\linewidth}p{0.29\linewidth}p{0.28\linewidth}p{0.18\linewidth}}
\toprule
Setting & Why it stresses allocation & Slice or route signal & Deployment status \\
\midrule
Hard GSM8K rationale & Answer checking saturates, but rationale validity
requires complementary LLM judgments. & Candidate generator and verifier
agreement. & Available before final audit label. \\
MBPP public-overfit & A cheap hidden-test verifier can dominate some LLM
signals but not all code-audit cases. & Public-test pass/fail and verifier
agreement. & Available before final hidden-test label. \\
JailbreakBench & Safety judges have conditional value on unsafe and
classifier-disagreement regions. & Classifier/disagreement proxy slices; human
safety label is audit-only.
& Deployable only for proxy slices, not for human-label slices.
\\
LLMBar & Preference failures differ across natural and adversarial subsets,
making specialist routing central. & Natural, adversarial instruction,
adversarial output, and neighbor subsets. & Dataset metadata available before routing. \\
RewardBench / Arena100K & Broad preference comparisons test whether stopped
panels should give way to full-call aggregation. & Preference-source and
candidate-pair metadata. & Dataset metadata available before routing. \\
SummEval & Scalar summary judging tests whether additional judges improve a
continuous audit target. & Summary dimension and judge-confidence proxy. &
Dimension metadata available; confidence is judge-derived. \\
MATH-500 & Difficult math checks whether broad ensembles remain useful beyond
cheap stopped panels. & Problem level and generator family. & Available as metadata. \\
HumanEval / GSM8K & Saturated verifier cases test whether the method refuses
unnecessary expansion. & Unit-test or answer-verifier result. & Verifier output available before routing. \\
\bottomrule
\end{tabular}
\caption{Experimental matrix. Each setting is included because it exercises a
different deployment decision: add complements, route specialists, stop early,
drop copies, or accept a full-call boundary. Human labels may define audit
slices for analysis, but only metadata, verifier outputs, classifier outputs,
or judge-disagreement proxies are deployable route signals.}
\label{tab:experimental-matrix}
\end{table}

The evaluation metric is held-out squared calibration risk for all tasks and
accuracy where labels are binary. Risk is the primary metric because the method
selects judges through calibrated conditional gain; accuracy is reported to
make the results legible for standard correctness, safety, and preference
audits \citep{guo2017calibration}. Average cost and average number of judge
calls are reported because the paper's object is a deployment policy rather
than an unconstrained aggregator.

The datasets are chosen to prevent a single story from explaining every result.
Hard GSM8K rationale and MBPP public-overfit test whether a cheap verifier and
a few LLM judges can be combined without defaulting to all calls. LLMBar tests
deployable conditional routing on natural and adversarial subsets, while JBB
tests whether safety value is concentrated on deployable classifier or
disagreement proxies and on human-label audit strata. Human labels are used
only for audit evaluation, not for deployment-time routing. Arena100K and
SummEval test stopping in non-saturated settings where extra judges can worsen
calibration. RewardBench and MATH-500 are boundary
cases where broad aggregation can remain attractive. HumanEval and ordinary
GSM8K are saturated sanity checks: after an objective verifier solves the audit
target, the correct policy action is to stop.

The baselines are similarly separated by deployment question. The flat panel
and full-call stacking baselines answer ``what if we call every judge?'' and
therefore form strong risk endpoints at high cost. Matched-size non-role panels
answer whether ordinary quality or correlation diversity can match the same
call budget without role conditioning. Reliability jury answers whether global
judge trustworthiness is enough. Frugal cascade answers whether a single
quality order with an uncertainty trigger is enough. Role allocation should win
only when the missing ingredient is conditional value relative to the current
panel or slice.

\section{Results}

The evidence chain follows the deployment actions induced by the role profile.
Table~\ref{tab:main-results} asks which call plan each setting supports. Tables
\ref{tab:strong-baselines} and \ref{tab:mechanisms} then separate the regimes:
where conditional specialists should be routed, where a cheap stopped panel is
enough, where copied signals should be dropped, and where the right endpoint is
still a broad full-call ensemble.

\begin{table}[t]
\centering
\scriptsize
\begin{tabular}{lrrrrrrrr}
\toprule
Dataset & \multicolumn{2}{c}{Single best} & \multicolumn{2}{c}{Flat all} &
\multicolumn{4}{c}{Role routed stop} \\
\cmidrule(lr){2-3}\cmidrule(lr){4-5}\cmidrule(lr){6-9}
& Risk & Acc. & Risk & Acc. & Risk & Acc. & Cost & Judges \\
\midrule
Hard GSM8K rationale & 0.2350 & 0.6253 & 0.2106 & 0.6670 & 0.2137 & 0.6843 & 2.90 & 2.90 \\
MBPP public-overfit & 0.0226 & 0.9767 & 0.0158 & 0.9617 & 0.0097 & 0.9900 & 1.52 & 1.70 \\
JBB-7 & 0.1183 & 0.8349 & 0.1291 & 0.8409 & 0.1094 & 0.8527 & 2.29 & 2.29 \\
LLMBar-7 & 0.2180 & 0.6822 & 0.2118 & 0.6692 & 0.1884 & 0.7334 & 3.46 & 3.46 \\
RewardBench-7 & 0.0308 & 0.9678 & 0.0280 & 0.9615 & 0.0291 & 0.9660 & 1.80 & 1.80 \\
Arena100K-7 & 0.2321 & 0.6257 & 0.2462 & 0.6186 & 0.2321 & 0.6257 & 1.00 & 1.00 \\
SummEval-7 scalar & 0.0450 & -- & 0.0601 & -- & 0.0450 & -- & 1.00 & 1.00 \\
MATH-500-5 & 0.0731 & 0.9167 & 0.0537 & 0.9309 & 0.0678 & 0.9202 & 1.70 & 1.70 \\
\bottomrule
\end{tabular}
\caption{Main held-out policy comparison across hard reasoning audits, code
overfit audits, safety, pairwise preference, reward modeling, and scalar
summarization. Role policies expose few-judge complement panels, one-step
stopping, specialist routing, and broad-ensemble endpoints.}
\label{tab:main-results}
\end{table}

Table~\ref{tab:main-results} translates held-out metrics into deployment
decisions. On hard GSM8K rationales, MBPP overfit, safety audit/proxy slices,
and LLMBar, role policies recover useful accuracy with fewer than a flat
panel's calls. Arena100K and SummEval produce a different action: keep the
strong single judge because expansion adds little value. RewardBench and
MATH-500 expose the full-panel endpoint, where extra broad signals can be worth
their cost when the researcher wants the lowest risk.

The complement regimes show why conditioning matters. In hard GSM8K rationale
audits, ordinary answer checking is not the target: the policy must decide
whether the reasoning is valid. The stopped role policy improves accuracy over
both the single-best and flat-all panels while invoking about three judges. In
MBPP public-overfit, the hidden-test verifier is cheap and strong, but it does
not eliminate all residual audit uncertainty. The role policy reaches the best
reported accuracy at a cost close to one and a half calls, illustrating the
intended combination of verifier-first stopping with selective LLM additions.

\begin{table}[t]
\centering
\scriptsize
\setlength{\tabcolsep}{4.2pt}
\begin{tabular}{lrrrrrr}
\toprule
Setting & \multicolumn{2}{c}{Best full-call} &
\multicolumn{2}{c}{\begin{tabular}{@{}c@{}}Best matched\\non-role\end{tabular}} &
\multicolumn{2}{c}{Role policy} \\
\cmidrule(lr){2-3}\cmidrule(lr){4-5}\cmidrule(lr){6-7}
& Risk & Cost & Risk & Cost & Risk & Cost \\
\midrule
Hard GSM8K rationale & 0.1963 & 6.10 & 0.2114 & 2.90 & 0.2137 & 2.90 \\
MBPP public-overfit & 0.0067 & 6.10 & 0.0117 & 1.61 & 0.0097 & 1.52 \\
JBB-7 DeepSeek & 0.1069 & 7.00 & 0.1151 & 1.90 & 0.1094 & 2.29 \\
LLMBar-7 DeepSeek & 0.1804 & 7.00 & 0.1967 & 2.50 & 0.1884 & 3.46 \\
LLMBar-7 Qwen3 & 0.2034 & 7.00 & 0.2190 & 2.20 & 0.2033 & 3.28 \\
LLMBar-7 JudgeLM & 0.1999 & 7.00 & 0.2215 & 2.30 & 0.2040 & 3.48 \\
RewardBench-7 DeepSeek & 0.0201 & 7.00 & 0.0284 & 1.50 & 0.0291 & 1.80 \\
Arena100K-7 DeepSeek & 0.2286 & 7.00 & 0.2321 & 1.00 & 0.2321 & 1.00 \\
SummEval-7 DeepSeek & 0.0446 & 7.00 & 0.0450 & 1.00 & 0.0450 & 1.00 \\
MATH-500-5 & 0.0536 & 5.00 & 0.0668 & 1.70 & 0.0678 & 1.70 \\
\bottomrule
\end{tabular}
\caption{Strong baseline comparison. Full-call aggregation can be the best
risk endpoint, but it invokes every judge. Role policies solve the deployment
problem of deciding when to buy a small stopped panel, when to route
specialists, and when to keep the full-call endpoint.}
\label{tab:strong-baselines}
\end{table}

Full-call stacking is a strong endpoint because it sees all judge outputs before
predicting. Role allocation answers the preceding operational question: which
outputs should be purchased in the first place? Table~\ref{tab:strong-baselines}
shows that role policies add the most value when information is conditional on
the current panel or route signal, as in safety and LLMBar. On hard GSM8K,
RewardBench, and MATH-500, matched panels or full-call endpoints can be just as
competitive. That is the intended regime-map reading: the policy tells the
researcher whether to buy conditional specialists, stop early, or pay for broad
aggregation.
The JBB proxy-routing audit in Table~\ref{tab:jbb-proxy-routing} isolates the
deployable safety case: routing on \texttt{gpt4\_cf}, not human labels, reaches
0.1094 risk at 2.29 calls versus the 7-call stack at 0.1069.

The matched-\(K\) comparison is the key judge-count fairness check. A top-\(k\)
panel can reuse the same number of judges, but its realized call cost can differ;
it selects judges by standalone validation
quality rather than conditional value. Correlation-diverse and quality-diverse
panels also spend a similar budget, but their notion of diversity is nominal or
pairwise rather than target-conditional. The gains on MBPP, the safety
audit/proxy setting, and the three LLMBar anchors show what the role profile
adds: it spends the same budget on judges whose residual information is useful
for the current panel and target slice \citep{kuncheva2003diversity}.

\begin{table}[t]
\centering
\scriptsize
\setlength{\tabcolsep}{3.7pt}
\begin{tabular}{lrrrrrr}
\toprule
Setting & \multicolumn{2}{c}{Reliability jury} &
\multicolumn{2}{c}{Frugal cascade} &
\multicolumn{2}{c}{Role policy} \\
\cmidrule(lr){2-3}\cmidrule(lr){4-5}\cmidrule(lr){6-7}
& Risk & Cost & Risk & Cost & Risk & Cost \\
\midrule
Hard GSM8K rationale & 0.1957 & 6.10 & 0.2182 & 2.76 & 0.2137 & 2.90 \\
MBPP public-overfit & 0.0059 & 6.10 & 0.0225 & 1.06 & 0.0097 & 1.52 \\
JBB-7 DeepSeek & 0.1382 & 7.00 & 0.1213 & 1.43 & 0.1094 & 2.29 \\
LLMBar-7 DeepSeek & 0.2058 & 7.00 & 0.2107 & 2.52 & 0.1884 & 3.46 \\
LLMBar-7 JudgeLM & 0.2113 & 7.00 & 0.2337 & 2.60 & 0.2040 & 3.48 \\
LLMBar-7 Qwen3 & 0.2232 & 7.00 & 0.2333 & 1.94 & 0.2033 & 3.28 \\
RewardBench-7 DeepSeek & 0.0268 & 7.00 & 0.0308 & 1.00 & 0.0291 & 1.80 \\
MATH-500-5 & 0.0631 & 5.00 & 0.0658 & 1.67 & 0.0678 & 1.70 \\
\bottomrule
\end{tabular}
\caption{SOTA-style allocation baselines. Reliability jury is full-call
multi-annotator aggregation; frugal cascade is confidence-triggered budgeted
routing. Role policies are most informative when useful judges are
slice-conditional, as in deployable LLMBar slices and safety proxy/audit
slices.}
\label{tab:sota-style-baselines}
\end{table}

Reliability jury estimates which judges are globally trustworthy, but does not
decide that a judge should be called only on a slice. Frugal cascade decides
when to call another globally ordered judge based on uncertainty, but it does
not model specialist roles. This explains Table~\ref{tab:sota-style-baselines}:
the baselines are strong in broad-complement regimes, while role routing is the
natural deployment action when failure modes are conditional.

These comparisons reveal two useful operating modes. In broad-complement
settings such as hard GSM8K and MBPP, full-call reliability juries can be the
lowest-risk endpoints because every signal contributes to the aggregate. In
slice-conditional settings such as safety proxy/audit slices and LLMBar,
useful information concentrates on failure modes. There the role policy is both
cheaper than the full-call jury and lower risk than the cascade.

\begin{figure}[t]
\centering
\includegraphics[width=0.92\linewidth]{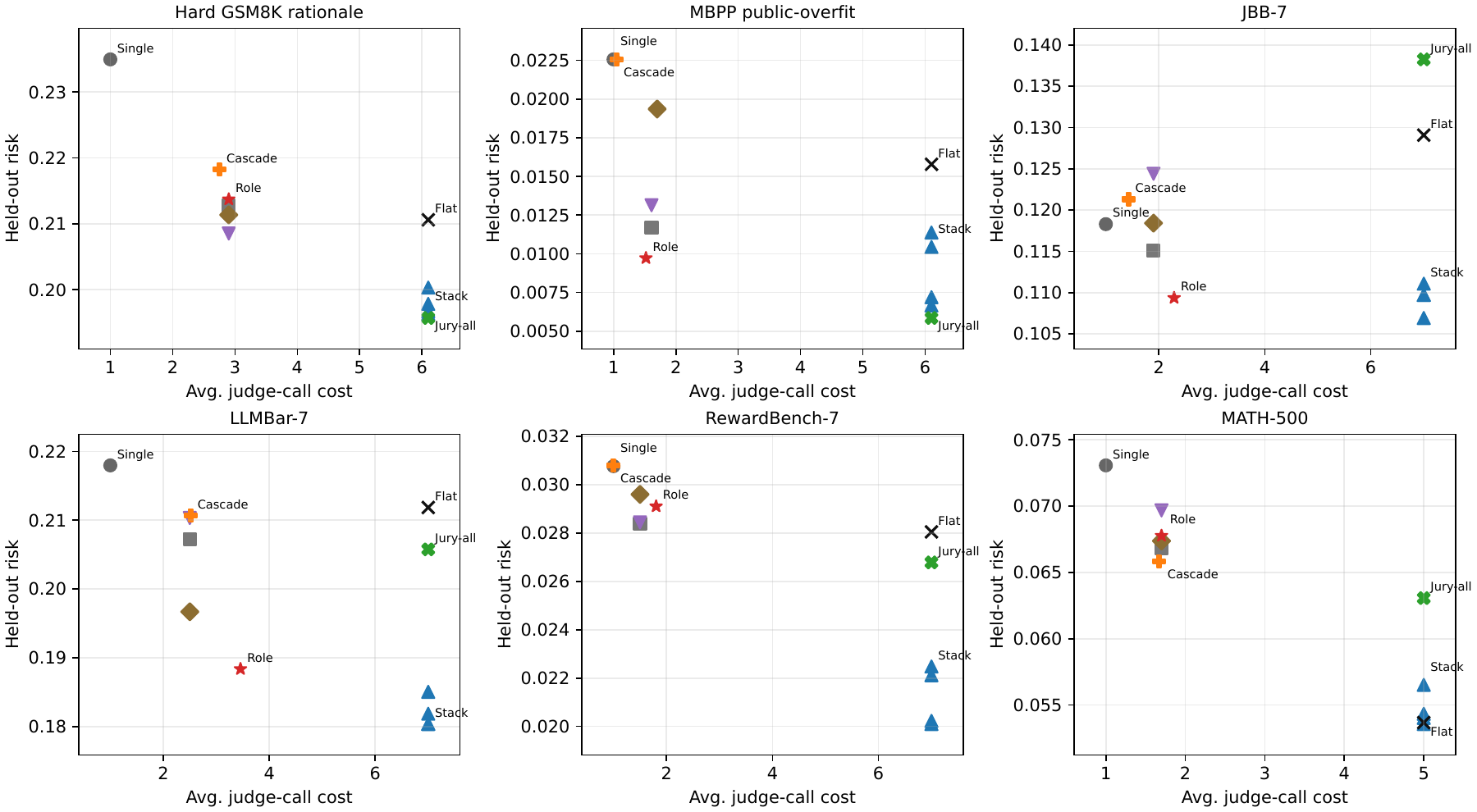}
\caption{Risk-cost frontier across representative settings. Each point is a
held-out policy evaluation averaged over 10 splits. Full-call stacking and
full-call jury can be low-risk endpoints in broad-ensemble regimes, but require
invoking every judge. Role policies occupy useful frontier regions when
specialists or cheap verifiers matter.}
\label{fig:risk-cost-frontier}
\end{figure}

\begin{table}[t]
\centering
\scriptsize
\setlength{\tabcolsep}{3.5pt}
\begin{tabular}{p{0.20\linewidth}p{0.31\linewidth}p{0.39\linewidth}}
\toprule
Policy action & Evidence & Interpretation \\
\midrule
Route specialists & LLMBar improves from flat-all risk 0.2118/accuracy 0.6692
to role risk 0.1884/accuracy 0.7334 at 3.46 calls; repeated routes appear on
adversarial and natural subsets across DeepSeek, Qwen3, and JudgeLM anchors. &
Useful judges are conditional on declared slices, not merely next in a global
quality order. \\
\midrule
Stop & Increasing \(\tau\) reduces calls on hard GSM8K (3.10 to 1.30), MBPP
(1.80 to 1.10), JBB (3.68 to 1.00), and MATH-500 (2.70 to 1.00); HumanEval and
ordinary GSM8K stop after the verifier or strong single judge. & The method
produces a practical validation-based stopping report for saturated targets. \\
\midrule
Drop copies & Adding four exact copies to LLMBar and JBB leaves role risk/cost
unchanged (LLMBar 0.1884/3.46; JBB 0.1094/2.29), while full-call jury cost
rises to 11 and risk worsens (LLMBar 0.2860; JBB 0.1594). & Conditional gain
identifies redundant signals even when nominal panel size and model count grow.
\\
\midrule
Expose boundaries & RewardBench and MATH-500 role policies are cheaper, but
full-call stacking gives lower risk (0.0201 vs 0.0291; 0.0536 vs 0.0678). &
The method is a regime detector: when broad ensemble information remains
valuable and cost is acceptable, call the full panel. \\
\bottomrule
\end{tabular}
\caption{Mechanism evidence for the four actions induced by the taxonomy:
route, stop, drop, and select broad-ensemble endpoints.}
\label{tab:mechanisms}
\end{table}

\paragraph{Routed specialists.}
Table~\ref{tab:routing} checks that routing is not merely an expensive global
panel in disguise. On LLMBar, the policy repeatedly sends different judges to
declared adversarial and natural subsets across three anchors. This is the
mechanism missing from global reliability juries and confidence cascades: a
judge can fail to clear the broad threshold while still clearing a slice
threshold.

\begin{table}[t]
\centering
\small
\begin{tabular}{llrr}
\toprule
Anchor & Slice & Routed judge & Frequency \\
\midrule
DeepSeek & adversarial\_gptinst & llama3\_8b\_v & 6/10 \\
DeepSeek & adversarial\_neighbor & llama3\_8b\_v & 5/10 \\
DeepSeek & natural & gemma3\_12b\_v & 6/10 \\
Qwen3 & adversarial\_gptout & gemma3\_12b\_v & 5/10 \\
Qwen3 & adversarial\_neighbor & mistral\_7b\_v & 6/10 \\
Qwen3 & natural & gemma3\_12b\_v & 5/10 \\
JudgeLM & adversarial\_gptout & gemma3\_12b\_v & 5/10 \\
JudgeLM & adversarial\_neighbor & mistral\_7b\_v & 6/10 \\
JudgeLM & natural & gemma3\_12b\_v & 5/10 \\
\bottomrule
\end{tabular}
\caption{Representative routed specialists on LLMBar. Frequencies count how
often a judge-slice route appears across 10 random splits.}
\label{tab:routing}
\end{table}

\paragraph{Stopping thresholds.}
Table~\ref{tab:threshold} varies the validation threshold. The pattern is not
that one threshold is universally best; it is that \(\tau\) gives the researcher
a transparent risk-cost dial. Higher thresholds save calls on hard GSM8K, MBPP,
JBB, and MATH-500. LLMBar is a useful exception: a more conservative threshold
also improves risk by avoiding sparse or redundant expansions.

\begin{table}[t]
\centering
\scriptsize
\begin{tabular}{lrrrr}
\toprule
Dataset and threshold & Risk & Acc. & Cost & Judges \\
\midrule
Hard GSM8K rationale, \(\tau=0.001\) & 0.2129 & 0.6790 & 3.10 & 3.10 \\
Hard GSM8K rationale, \(\tau=0.005\) & 0.2137 & 0.6843 & 2.90 & 2.90 \\
Hard GSM8K rationale, \(\tau=0.020\) & 0.2318 & 0.6373 & 1.30 & 1.30 \\
\midrule
MBPP public-overfit, \(\tau=0.001\) & 0.0078 & 0.9920 & 1.53 & 1.80 \\
MBPP public-overfit, \(\tau=0.005\) & 0.0097 & 0.9900 & 1.52 & 1.70 \\
MBPP public-overfit, \(\tau=0.020\) & 0.0206 & 0.9787 & 1.01 & 1.10 \\
\midrule
JBB, \(\tau=0.001\) & 0.1078 & 0.8688 & 3.68 & 3.68 \\
JBB, \(\tau=0.005\) & 0.1094 & 0.8527 & 2.29 & 2.29 \\
JBB, \(\tau=0.020\) & 0.1183 & 0.8349 & 1.00 & 1.00 \\
\midrule
LLMBar, \(\tau=0.001\) & 0.1954 & 0.7303 & 4.10 & 4.10 \\
LLMBar, \(\tau=0.005\) & 0.1884 & 0.7334 & 3.46 & 3.46 \\
LLMBar, \(\tau=0.020\) & 0.1834 & 0.7443 & 2.00 & 2.00 \\
\midrule
MATH-500, \(\tau=0.001\) & 0.0617 & 0.9209 & 2.70 & 2.70 \\
MATH-500, \(\tau=0.005\) & 0.0678 & 0.9202 & 1.70 & 1.70 \\
MATH-500, \(\tau=0.020\) & 0.0731 & 0.9167 & 1.00 & 1.00 \\
\bottomrule
\end{tabular}
\caption{Threshold sensitivity for role-routed stopping. Conservative
thresholds reduce calls and provide an explicit stopping condition: add no
remaining judge whose validation gain is below \(\tau\).}
\label{tab:threshold}
\end{table}

\paragraph{Copy stress test.}
The copy role should change deployment, not merely interpretation. We therefore
add four exact copies of an existing DeepSeek judge to LLMBar and JBB. The role
policy is unchanged because the copies have zero conditional validation gain
after the original signal is present. Full-call baselines still pay for the
copies, and reliability jury becomes worse because duplicated votes are
overweighted.

\begin{table}[t]
\centering
\scriptsize
\begin{tabular}{llrrrrrrrr}
\toprule
Setting & Condition & \multicolumn{2}{c}{Flat all} &
\multicolumn{2}{c}{Jury all} & \multicolumn{2}{c}{Cascade} &
\multicolumn{2}{c}{Role} \\
\cmidrule(lr){3-4}\cmidrule(lr){5-6}\cmidrule(lr){7-8}\cmidrule(lr){9-10}
& & Risk & Cost & Risk & Cost & Risk & Cost & Risk & Cost \\
\midrule
LLMBar & base & 0.2118 & 7.00 & 0.2058 & 7.00 & 0.2107 & 2.52 & 0.1884 & 3.46 \\
LLMBar & +4 copies & 0.2118 & 11.00 & 0.2860 & 11.00 & 0.2107 & 3.80 & 0.1884 & 3.46 \\
JBB & base & 0.1291 & 7.00 & 0.1382 & 7.00 & 0.1213 & 1.43 & 0.1094 & 2.29 \\
JBB & +4 copies & 0.1291 & 11.00 & 0.1594 & 11.00 & 0.1213 & 1.65 & 0.1094 & 2.29 \\
\bottomrule
\end{tabular}
\caption{Redundant-copy stress test. Four exact copies of an existing judge
increase the apparent pool size, but role-conditioned stopping ignores them
after their conditional gain vanishes.}
\label{tab:copy-stress}
\end{table}

Taken together, the results support a policy interpretation of judge diversity.
When a new judge supplies broad conditional information, it should enter the
global panel. When its value is concentrated on a declared slice, it should be
routed rather than called everywhere. When its gain vanishes after conditioning
on the current panel, it should be dropped even if it increases nominal model
diversity. When full-call aggregation remains lower risk and the cost is
acceptable, the regime map marks the full panel as the deployment endpoint.

\section{Related Work}

\paragraph{LLM-as-a-judge evaluation.}
LLM judges are widely used for open-ended generation, instruction following,
translation quality, preference comparison, and rubric scoring
\citep{zheng2023judging,liu2023geval,kocmi2023state,zhu2023judgelm,
kim2024prometheus2,dubois2024length}. This line of work establishes that LLMs
can be useful evaluators, but also documents evaluator-specific biases such as
position and length effects \citep{wang2024fair,dubois2024length}. Benchmarks
such as LLMBar, RewardBench, Arena100K, SummEval, MATH/MATH-500, GSM8K, MBPP,
and HumanEval define the targets and stress cases used in this paper
\citep{zeng2024llmbar,lambert2024rewardbench,chiang2024chatbot,
fabbri2021summeval,hendrycks2021math,lightman2023lets,cobbe2021training,
austin2021program,chen2021evaluating}. Our contribution is not a new judge
benchmark. It is a policy for deciding which available judge signals to call
under a finite audit budget.

The same literature also motivates the need for conditional allocation rather
than one global judge ranking. LLMBar constructs adversarial preference cases
that expose evaluator failures \citep{zeng2024llmbar}; AlpacaEval and FairEval
document length and position biases in automatic judges
\citep{dubois2024length,wang2024fair}; and JailbreakBench separates safety
evaluation from ordinary helpfulness or preference evaluation
\citep{chao2024jailbreakbench}. These findings imply that a judge's usefulness
can depend sharply on the failure mode. Role-conditioned routing treats that
dependence as a deployment object: a judge may be worth calling on adversarial,
safety-proxy, or disagreement regions without being worth calling everywhere.

\paragraph{Judge panels and multi-agent evaluators.}
Several works study using more than one LLM evaluator, either as panels or
multi-agent discussions \citep{verga2024juries,chaneval2024}. These methods
motivate judge diversity, but they do not by themselves determine whether an
additional judge should be called globally, routed to a slice, or dropped as a
copy. Role-conditioned allocation treats panel construction as a conditional
value problem: the value of a candidate depends on the current panel, the
target distribution, the slice, and the cost.
Correlated-error audits show that nominal panel size can substantially overstate
effective information \citep{kohli2026nine}, while calibrated full-panel results
show that weak but nonredundant judges can remain useful when their signals are
learnable \citep{li2026calibrate}. Our policy reconciles these observations at
deployment time by retaining a signal only when its finite-sample conditional
gain justifies its call cost.

\paragraph{Annotator aggregation and preference models.}
Reliability-based aggregation has a long history in multi-annotator learning
\citep{dawid1979maximum,raykar2010learning,whitehill2009whose}, and pairwise
comparison models such as Bradley--Terry remain standard tools for preference
aggregation \citep{bradley1952rank}. These methods estimate global or
item-conditioned annotator reliability from observed labels. Our setting is
different because a deployment system must decide which judge outputs to
observe in the first place. A judge can be low-reliability globally but useful
on one slice, or high-reliability but redundant after a verifier enters the
panel.

\paragraph{Ensembles, deferral, cascades, and calibration.}
Stacked generalization and ensemble selection show how to combine many model
outputs once they are observed \citep{wolpert1992stacked,caruana2004ensemble}.
Classifier-ensemble work also studies diversity measures and their limits
\citep{kuncheva2003diversity}. Learning-to-defer methods train systems to
route examples to a human or expert when delegation improves task performance
\citep{madras2018predict,mozannar2020consistent}. Model cascades and routers
reduce inference cost by calling stronger models only when needed
\citep{chen2024frugalgpt,ong2025routellm}. More generally, wrapper selection,
conditional-redundancy criteria, and budgeted classifier cascades select
predictive signals under validation or acquisition costs
\citep{kohavi1997wrappers,brown2012conditional,chen2012cascade}; active feature
acquisition makes the corresponding per-example decision about which costly
features to observe \citep{shim2018joint}. Cascaded Selective Evaluation applies
confidence-based escalation to LLM judges with a target human-agreement guarantee
\citep{jung2025trust}. In contrast, our setting selects a set-valued panel policy
from finite audit data, allowing global complements, slice-routed specialists,
and unused conditional copies. Role-conditioned allocation applies
the same conditional-computation question to evaluation itself. Its units are
auditable judge calls: use calibrated validation gain
\citep{guo2017calibration} to decide whether an output should be obtained
globally, routed to a declared slice, or left uncalled. The LLM-judge setting
adds three constraints that ordinary routers do not address together:
route signals must be deployable before the routed judge call, human-label
strata can be audit diagnostics without being route inputs, and nominally
different prompts or models may be conditional copies after the current panel
has already been observed.

\section{Discussion}

Judge diversity becomes useful when it changes a deployment action. A model
family, prompt template, or reward head can look diverse on paper and still be
a copy after conditioning on the current panel; a weak standalone judge can be
exactly the call worth making on one failure slice. The practical workflow is a
calibration loop: collect a labeled audit set, run the candidate judge pool
once, declare decision-relevant slices, and fit the stopped role policy. The
output is a call plan with global judges, routed specialists, and unused judges
whose validation gain did not justify their cost.

That plan has a direct operational reading. If a deterministic verifier
dominates, keep it and stop. If broad gains plateau but slice gains remain,
route specialists. If copied signals appear, drop them without changing the
rest of the policy. If the full panel is still the lowest-risk endpoint and the
budget allows it, pay for that endpoint. Across repeated audit splits, risk-cost
stability and exact plan identity should be read separately: stable risk with
variable routes calls for frequency checking, keeping recurring calls and
collecting more audit labels before relying on low-frequency specialists. This
frequency audit is a deployment diagnostic, not a statistical guarantee.

The threshold is the risk-cost dial. Lower thresholds keep marginal calls when
evaluation errors are expensive; higher thresholds produce leaner policies when
latency or budget dominates. If small threshold changes alter the selected
panel, the audit set is signaling instability; if the same copy, route, or stop
decisions persist, the call plan is more credible for the next evaluation
batch.

\section{Deployment Extensions}

The same policy interface scales along three axes. Larger judge pools can keep
the role profile while replacing finite-cell means with smoothed, cross-fitted,
or parametric calibrators when joint output cells become sparse. Declared
slices can be extended by automatic slice discovery: discover candidate failure
regions, validate whether any judge has conditional value there, and route only
specialists that clear threshold. Normalized call costs can also be replaced by
actual API prices, latency, safety-review budget, or carbon budget; the output
remains a validation-backed plan of global calls, routed specialists, and
stopped candidates.

\section{Conclusion}

We proposed role-conditioned panel policies for LLM judge allocation. The shift
is from describing judge diversity to deciding judge calls. Target-relative
profiles identify copies, complements, and specialists; validation-stopped
construction turns them into global and routed policies; held-out evaluation
reports the resulting risk-cost tradeoff. The resulting regime map tells LLM
researchers when to drop, add, route, stop, or pay for a full panel.

The broader point is that a larger panel is not automatically more reliable,
and a smaller panel is not automatically more efficient if it drops conditional
information. Future judge-panel studies should therefore report not only which
evaluator scored best, but also which additional judges were worth calling,
where they were worth routing, and why panel construction stopped.

\section*{Acknowledgements}
This work was supported by the National Natural Science Foundation of China
(62372483).

\bibliographystyle{iclr2026_conference}
\bibliography{effective_judges_references}

\appendix

\section{Full Role Profile and Construction Details}

\subsection{Role Profiles With Specialization Ratios}

The main text uses the compact role profile
\(\Profile_{P,\F,S}(j)=(C_P,\{A_f\}_{f\in\F})\). In the experiments we also
track a specialization ratio
\[
\rho_f(j\mid S)
=
\frac{g_{P_f}(j\mid S)}{g_P(j\mid S)+\epsilon_0},
\]
where \(\epsilon_0>0\) prevents division by zero. The ratio is not used to make
complement and specialist mutually exclusive. Instead, it is a diagnostic for
concentration: a judge may have positive broad gain and still be unusually
valuable on one declared slice. This is why the role table in the main text
allows a complement-plus-specialist role.

The role interpretation used throughout the experiments is:
\begin{itemize}[leftmargin=*]
\item \textbf{Copy:} broad gain and all slice gains fall below the declared
threshold after conditioning on the current panel.
\item \textbf{Broad complement:} cost-adjusted broad validation gain exceeds
\(\tau_P\), so the judge is added to the global panel.
\item \textbf{Slice specialist:} cost-adjusted slice validation gain exceeds
\(\tau_f\), so the judge is invoked only on examples routed to that slice.
\item \textbf{Complement plus specialist:} broad gain is positive and at least
one slice gain is concentrated; the judge can enter globally and can also be
prioritized for interpretation on that slice.
\end{itemize}

\subsection{Construction Algorithm}

The complete construction procedure is:
\begin{enumerate}[leftmargin=*]
\item Collect a labeled audit set and run all candidate judges on it.
\item Declare slices \(\F\) that are meaningful for the target deployment
distribution.
\item Split the audit set into construction-fit, construction-validation, and
final-test portions.
\item Start from an empty global panel or a user-specified seed panel.
\item Fit the pattern calibrator for the current global panel on the fit split.
\item Estimate each remaining candidate's broad validation gain on the
validation split.
\item Add the candidate with the largest positive cost-adjusted broad gain if
that gain exceeds \(\tau_P\); otherwise stop global construction.
\item For each slice, repeat the same greedy search after conditioning on the
selected global panel and any specialists already assigned to that slice.
\item Refit calibrators for the selected global and routed paths on the full
construction split.
\item Evaluate the resulting calling policy on final-test examples only.
\end{enumerate}

Judge outputs are canonicalized before pattern construction: binary and
preference labels are mapped to normalized symbols, confidence suffixes are
stripped for the pattern table, and numeric rubric scores such as SummEval's
1--5 judgments remain ordinal cell labels for the pattern calibrator. Numeric
scores are normalized to \([0,1]\) only for scalar stacking and cascade
baselines. Experiments use complete-case rows for the declared judge pool; rows
with missing or unparseable selected judge outputs are excluded before splitting.
When two candidates have the same cost-adjusted gain, ties are resolved by raw
gain and then by the judge identifier, making the construction deterministic
for a fixed split.

This procedure produces both a policy and a stopping report. The stopping report is the
set of failed inequalities: after stopping, every unused broad candidate is
below \(\tau_P\), and every unused routed candidate is below the corresponding
slice threshold. The stopping report is tied to the search space used by the audit.
With the default greedy search, it records that no single additional call is
justified under the finite audit set, threshold, and cost model. With beam,
pair-addition, or subset proposals, the same reporting format certifies the
expanded candidate moves.

\section{Dataset and Slice Details}

\paragraph{Hard GSM8K rationale audits.}
Ordinary GSM8K answer checking is too easy for the main claim because an answer
verifier can saturate the target. The hard rationale setting instead asks
whether a candidate solution rationale is valid. This creates a complement
regime: the verifier is cheap and useful, but LLM judges can still provide
conditional information about reasoning validity.

\paragraph{MBPP public-test overfit audits.}
The code setting asks whether a candidate program has overfit public tests or
generalizes to hidden tests. The hidden-test verifier is cheap and often
dominant, but LLM judges may still help on residual code-audit cases. This
setting tests whether the policy can combine a deterministic verifier with a
small number of complementary LLM calls instead of always invoking the full
panel.

\paragraph{JailbreakBench safety.}
The safety setting includes human-labeled safe and unsafe responses. Slices
include human-label strata for audit analysis and deployable proxy regions
defined by safety-classifier output or judge disagreement. Human labels are not
available at deployment time, so they serve as audit strata rather than route
inputs. The practical deployment question is whether a safety judge should be
invoked globally or only on proxy regions where the current panel is
unreliable; label-conditioned JBB tables report audit-slice evidence, while
classifier and disagreement proxies define deployable routes.

\begin{table}[t]
\centering
\scriptsize
\setlength{\tabcolsep}{3.8pt}
\begin{tabular}{lrrrp{0.29\linewidth}}
\toprule
Policy & Risk (95\% CI) & Acc. & Cost & Deployment reading \\
\midrule
Single best & \(0.1183 \pm 0.0105\) & 0.8349 & 1.00 &
Cheapest one-call reference. \\
Flat all & \(0.1291 \pm 0.0052\) & 0.8409 & 7.00 &
All safety judges on every item. \\
Frugal cascade & \(0.1213 \pm 0.0115\) & 0.8376 & 1.43 &
Uncertainty-triggered global order. \\
Full-call stack & \(0.1069 \pm 0.0034\) & 0.8450 & 7.00 &
Best full-call risk endpoint. \\
Role global stop & \(0.1199 \pm 0.0098\) & 0.8258 & 1.90 &
Stopped panel without proxy routing. \\
Role routed stop & \(0.1094 \pm 0.0106\) & 0.8527 & 2.29 &
Specialists routed on \texttt{gpt4\_cf} proxy slices. \\
\bottomrule
\end{tabular}
\caption{JailbreakBench safety proxy-routing audit. The route signal is the
dataset's GPT-4 classifier field \texttt{gpt4\_cf}
\citep{chao2024jailbreakbench,openai2023gpt4}, which is stored separately from
human-majority labels and is used here as a deployable proxy slice signal.
Across 10 splits, the routed policy selected extra specialists on the
classifier-safe proxy slice in 4 splits and on the classifier-unsafe proxy
slice in 2 splits.}
\label{tab:jbb-proxy-routing}
\end{table}

This audit is the safety version of the paper's central deployment question.
The full-call stack remains the best risk endpoint because it sees all seven
judge outputs, but role routing nearly reaches that endpoint while buying about
one third of the calls. The selected specialists are not fixed globally:
\texttt{mistral\_7b\_safety}, \texttt{deepseek\_v4\_flash\_safety},
\texttt{prometheus\_7b\_safety}, and \texttt{selene\_8b\_safety} appear on
different \texttt{gpt4\_cf} proxy slices across splits. This is the desired
behavior for a live safety audit: use a cheap classifier proxy to decide where
the panel needs extra scrutiny, and keep the stopped global panel elsewhere.

\paragraph{LLMBar preference.}
LLMBar is the main specialist-routing benchmark. Its natural and adversarial
subsets induce different failure modes. The repeated routes in the main text
show that the policy selects different judges for adversarial instruction,
adversarial output, adversarial neighbor, and natural subsets, rather than
expanding the full panel uniformly.

\paragraph{RewardBench and Arena100K preference.}
These preference settings test broad-ensemble behavior. Role policies expose
cheap frontier points, and full-call aggregation can remain the lowest-risk
endpoint. The regime map identifies when to pay for a broad ensemble and when a
stopped policy is already sufficient.

\paragraph{SummEval scalar judging.}
SummEval evaluates scalar summary quality. It is useful because additional
judges can worsen or barely improve the risk-cost tradeoff. In this regime, a
one-step stopped policy is a meaningful outcome rather than a failed panel. The
pattern policy treats each 1--5 rubric output as a discrete cell symbol, while
scalar baselines use the normalized score value.

\paragraph{MATH-500 correctness.}
MATH-500 tests whether hard math correctness benefits from broad ensembles.
The stopped role policy gives a cheaper point, while full-call stacking can be
lower risk. This supports the regime-map framing.

\paragraph{HumanEval and ordinary GSM8K.}
These are saturated stopping checks. If a unit-test verifier or answer verifier
already solves the audit target, the correct allocation decision is to stop
rather than to claim artificial panel gains.

\section{Judge Pool Disclosure}
\label{app:judge-pools}

Table~\ref{tab:judge-pools} lists the candidate signals used by the main
experiments. Names with suffixes \texttt{\_v}, \texttt{\_s},
\texttt{\_safety}, or task-specific correctness suffixes are direct-schema
judge outputs. Verifiers have normalized cost \(0.1\); all LLM judge calls have
normalized cost \(1.0\). Route keys are treated as pre-available metadata,
verifier outputs, classifier outputs, or already-observed proxy signals. If a
deployment must call an additional model to compute a route key, that call
should be added to the cost model before refitting the policy.
The candidate model families are DeepSeek, Qwen2.5/Qwen3, Gemma, Llama,
Mistral, Prometheus, JudgeLM, and Selene
\citep{deepseekai2024v3,qwen2024qwen25,qwen2025qwen3,
gemmateam2025gemma3,grattafiori2024llama3,jiang2023mistral,
kim2024prometheus2,zhu2023judgelm,bogolin2025selene}.

\begin{table}[t]
\centering
\scriptsize
\setlength{\tabcolsep}{3.0pt}
\begin{tabular}{p{0.18\linewidth}p{0.46\linewidth}p{0.16\linewidth}p{0.11\linewidth}}
\toprule
Setting & Candidate signals & Route key & Cheap verifier \\
\midrule
Hard GSM8K rationale &
DeepSeek, Llama-3.1, Mistral, Prometheus, Qwen2.5, Selene rationale judges;
GSM8K answer verifier. & candidate kind & answer verifier \\
MBPP public-overfit &
DeepSeek, Llama-3.1, Mistral, Prometheus, Qwen2.5, Selene overfit judges;
hidden-unit verifier. & candidate kind & hidden-unit verifier \\
JBB-7 &
DeepSeek, Gemma-3, Llama-3.1, Mistral, Prometheus, Qwen2.5, Selene safety
judges. & \texttt{gpt4\_cf} & none \\
LLMBar-7 &
DeepSeek/Qwen3/JudgeLM anchor plus Gemma-3, Llama-3, Mistral, Prometheus,
Qwen2.5, and Selene preference judges. & subset & none \\
RewardBench / Arena100K / SummEval &
DeepSeek, Gemma-3, Llama-3, Mistral, Prometheus, Qwen2.5, and Selene
preference or scalar-summary judges. & subset or none & none \\
MATH-500 &
Llama-3.1, Mistral, Prometheus, Qwen2.5, and Selene math-correctness judges.
& candidate model & none \\
HumanEval / GSM8K answer &
Task verifier plus Llama-3.1, Mistral, Prometheus, Qwen2.5, Selene, and where
available DeepSeek/Gemma correctness judges. & candidate model or kind &
unit-test or answer verifier \\
\bottomrule
\end{tabular}
\caption{Judge-pool disclosure for the main experiments. The table reports the
signals available to the allocation policy before it selects global calls,
routed specialists, or stopped candidates.}
\label{tab:judge-pools}
\end{table}

Table~\ref{tab:complete-case} reports the complete-case filtering used before
random splitting. The main seven-judge panels drop at most two rows. The
near-duplicate prompt-variant audit has a higher drop rate because one
letter-prompt judge has many unparseable outputs; that audit is therefore read
as a complete-case prompt-variant stress test rather than as a claim about
parse robustness.

\begin{table}[t]
\centering
\scriptsize
\begin{tabular}{lrrrrl}
\toprule
Setting & Rows & Judges & Complete & Dropped & Main unparseable source \\
\midrule
LLMBar-7 & 838 & 7 & 837 & 1 & Selene 1 \\
JBB-7 & 300 & 7 & 298 & 2 & Prometheus 2 \\
MBPP public-overfit & 300 & 7 & 300 & 0 & -- \\
SummEval-7 & 1600 & 7 & 1600 & 0 & -- \\
Math/GSM8K correctness & 300 & 8 & 300 & 0 & -- \\
LLMBar prompt variants & 838 & 10 & 680 & 158 & Prometheus-letter 155 \\
\bottomrule
\end{tabular}
\caption{Complete-case filtering before policy construction. Rows with missing
or unparseable outputs for the declared judge pool are excluded before the
construction/final-test split.}
\label{tab:complete-case}
\end{table}

\section{Baseline Implementation Details}

All baseline choices use the same construction-validation split as the role
policy, and all reported numbers are computed on the final-test split only.
This matters because full-call aggregation and cascades have enough flexibility
to overfit a small audit if their regularization, order, or thresholds are
chosen after looking at final-test outcomes. We therefore treat baseline
selection as part of the deployment procedure rather than as an oracle
leaderboard.

\paragraph{Single best and flat all.}
The single-best baseline selects the judge with lowest validation risk and
invokes only that judge on final-test examples. The flat-all baseline invokes
every candidate judge for every example and calibrates on the joint output
pattern.

\paragraph{Matched-size non-role panels.}
Matched-size panels use the same average number of calls as the stopped role
policy but select judges without role conditioning. The top-\(k\) version uses
standalone validation quality. The correlation-diverse version discourages
highly correlated judge outputs. The quality-diverse version balances
standalone quality with nominal diversity. These baselines ask whether
copy/complement/specialist roles add value beyond ordinary diversity
heuristics. The matched budget is fixed from construction-validation behavior,
then evaluated once on final-test examples.

\paragraph{Full-call stacking.}
Full-call ridge and logistic stacking are supervised aggregation endpoints:
they observe every judge output before predicting. Pairwise variants include
features derived from pairwise judge-output interactions. These methods can be
excellent low-risk endpoints, but they answer a different question from the
allocation policy because their call cost is fixed at the full panel. The
regularized linear or logistic variant used for a setting is selected on
construction-validation risk, not on final-test risk.

\paragraph{Reliability jury.}
The reliability jury treats judges as noisy annotators and estimates
label-conditional error behavior from construction data. It is a strong
multi-annotator aggregation baseline, especially when judge reliability is
mostly global. It does not decide that a judge should be called only on one
slice.

\paragraph{Frugal cascade.}
The cascade orders judges by validation quality and invokes additional judges
when the current calibrated prediction is uncertain. It is a strong cost-aware
baseline when one global order is adequate. It differs from role routing
because it does not identify slice specialists that should be called only on
declared failure modes. Its uncertainty threshold is chosen on the
construction-validation split under the same normalized cost model as the role
policy.

\section{Split-Level Variation}
\label{app:split-variation}

Table~\ref{tab:split-ci} reports the split-level uncertainty behind the main
risk table. Each entry is the mean held-out risk over 10 random splits with a
95\% confidence interval computed as
\(\bar r \pm t_{0.975,9}\,s/\sqrt{10}\), where \(s\) is the sample standard
deviation across splits. These intervals measure random split variation, not
uncertainty over future datasets or changing judge models.

\begin{table}[t]
\centering
\scriptsize
\setlength{\tabcolsep}{4.2pt}
\begin{tabular}{lrrr}
\toprule
Setting & Single best risk & Flat-all risk & Role policy risk \\
\midrule
Hard GSM8K rationale & \(0.2350 \pm 0.0055\) & \(0.2106 \pm 0.0116\) & \(0.2137 \pm 0.0082\) \\
MBPP public-overfit & \(0.0226 \pm 0.0039\) & \(0.0158 \pm 0.0050\) & \(0.0097 \pm 0.0096\) \\
JBB-7 & \(0.1183 \pm 0.0105\) & \(0.1291 \pm 0.0052\) & \(0.1094 \pm 0.0106\) \\
LLMBar-7 & \(0.2180 \pm 0.0116\) & \(0.2118 \pm 0.0098\) & \(0.1884 \pm 0.0143\) \\
RewardBench-7 & \(0.0308 \pm 0.0015\) & \(0.0280 \pm 0.0019\) & \(0.0291 \pm 0.0019\) \\
Arena100K-7 & \(0.2321 \pm 0.0034\) & \(0.2462 \pm 0.0046\) & \(0.2321 \pm 0.0034\) \\
SummEval-7 scalar & \(0.0450 \pm 0.0008\) & \(0.0601 \pm 0.0022\) & \(0.0450 \pm 0.0008\) \\
MATH-500-5 & \(0.0731 \pm 0.0033\) & \(0.0537 \pm 0.0049\) & \(0.0678 \pm 0.0058\) \\
\bottomrule
\end{tabular}
\caption{Split-level 95\% confidence intervals for the main held-out risk
comparisons. The role column uses the routed policy when routing is selected
and the global stopped policy in one-step stopping regimes.}
\label{tab:split-ci}
\end{table}

\section{Additional Interpretation of Main Results}

\paragraph{Few-judge complement regimes.}
Hard GSM8K rationale checking is neither saturated answer verification nor pure
slice routing. The single-best judge reaches 0.6253 accuracy, the flat panel
reaches 0.6670, and the stopped role policy reaches 0.6843 with about 2.9
calls. The best full-call and reliability-jury endpoints can be lower risk, but
the role policy recovers much of the benefit without paying for every judge.

MBPP public-overfit is easier but still non-saturated. The role policy reaches
0.9900 accuracy at cost 1.52, exceeding flat-all accuracy while using far fewer
calls. The result illustrates a practical pattern: a cheap verifier can be a
dominant signal while a small number of LLM judges remain useful.

\paragraph{Specialist-routing regimes.}
LLMBar is the clearest case where the taxonomy becomes a calling policy.
Routing slice specialists improves accuracy from 0.6692 for flat all and
0.6822 for single best to 0.7334 at 3.46 calls. The same qualitative pattern
appears under Qwen3 and JudgeLM anchors. JBB shows a related safety pattern:
the role policy is near the best full-call risk endpoint while using roughly a
third of the full-panel cost on proxy/audit slices. Only proxy slices based on
classifier outputs or judge disagreement are valid deployment-time route
signals; human-label strata are audit diagnostics.

\paragraph{One-step stopping regimes.}
Arena100K and SummEval demonstrate one-step stopping in non-saturated settings.
Expanding the panel worsens or barely improves the risk-cost tradeoff, so the
policy keeps a strong single judge. HumanEval and ordinary GSM8K are stronger
sanity checks: once a verifier solves the target, all remaining LLM judges have
zero useful validation gain.

\paragraph{Broad-ensemble boundary regimes.}
RewardBench and MATH-500 are boundary cases. The stopped role policies are
cheaper, but full-call stacking can remain lower risk. This is the intended
regime diagnosis: if the target still benefits from broad ensemble information
and cost is acceptable, the policy tells the researcher to keep the full panel.

\section{Deployment Robustness Audits}
\label{app:deployment-robustness}

After the initial risk-cost frontier is known, the deployment owner should run
three compact checks: whether selected pattern tables are sparse, whether the
call plan is stable with fewer audit labels, and whether the expected slice mix
matches the construction audit set. The owner then applies the relevant cost
model.
Because our deployable calibrator is a finite pattern table, we first audit how
often the selected call plan sees a joint judge pattern absent from its fitting
split and therefore falls back to the split mean. Table~\ref{tab:pattern-sparsity}
shows that the sparse-pattern pressure is concentrated in routed LLMBar/JBB
specialists; MBPP, MATH-500, and SummEval stop at short paths and have essentially
no fallback. The diagnostic is operational: if fallback is high on a target
slice, collect more slice labels or cap route depth before deploying that route.

\begin{table}[t]
\centering
\scriptsize
\setlength{\tabcolsep}{3.0pt}
\begin{tabular}{lrrrrrr}
\toprule
Setting & Fit/Val/Test & Calls & Global cells & Max route cells
& Val fallback & Test fallback \\
\midrule
LLMBar-7 & 209/209/419 & 3.46 & 8.3 & 23.7 & 8.7\% & 4.4\% \\
JBB-7 & 74/75/149 & 2.29 & 4.8 & 5.0 & 4.3\% & 2.2\% \\
MBPP public-overfit & 75/75/150 & 1.70 & 3.4 & 0.0 & 0.0\% & 0.0\% \\
MATH-500 & 200/201/401 & 1.70 & 3.4 & 0.0 & 0.2\% & 0.1\% \\
SummEval-7 & 400/400/800 & 1.00 & 5.0 & 0.0 & 0.0\% & 0.0\% \\
\bottomrule
\end{tabular}
\caption{Pattern-table sparsity for the selected role policy. Each row averages
10 random construction/test splits. Cells are occupied response-pattern cells in
the refit construction calibrator; route cells report the largest selected
slice-specific table in the split. Fallback is the fraction of validation or
final-test examples whose invoked pattern was unseen in the corresponding fit
table.}
\label{tab:pattern-sparsity}
\end{table}

Table~\ref{tab:audit-size-stability} fixes the held-out test split and varies
the fraction of the construction audit set used for fitting and validation. The
half-audit policies already recover the main LLMBar and JBB proxy risk-cost
behavior, while exact call-plan identity is less stable. This is a useful
diagnostic: the policy can be deployed when risk and cost are stable, and more
audit labels should be collected when the exact route set matters.

\begin{table}[t]
\centering
\scriptsize
\begin{tabular}{llrrrr}
\toprule
Setting & Audit fraction & Risk & Accuracy & Cost & Plan Jaccard \\
\midrule
LLMBar & 50\% & 0.1904 & 0.7371 & 3.07 & 0.246 \\
LLMBar & 100\% & 0.1884 & 0.7334 & 3.46 & 1.000 \\
JBB proxy & 50\% & 0.1066 & 0.8570 & 1.98 & 0.238 \\
JBB proxy & 100\% & 0.1094 & 0.8527 & 2.29 & 1.000 \\
MBPP public-overfit & 50\% & 0.0306 & 0.9680 & 0.65 & 0.083 \\
MBPP public-overfit & 100\% & 0.0097 & 0.9900 & 1.52 & 1.000 \\
\bottomrule
\end{tabular}
\caption{Audit-size stability. The held-out test split is fixed for each seed;
the construction audit set is reduced before fitting the role policy. Plan
Jaccard compares the global and routed call set with the full-audit plan for
the same split. Entries are means over 10 random splits.}
\label{tab:audit-size-stability}
\end{table}

Table~\ref{tab:slice-shift} changes the next-batch slice mix while fitting the
call plan on a different construction mix. This is the deployment reading:
route policies remain useful when the construction audit contains enough
examples from the deployment slice, and the table tells the owner when to buy
more slice labels before trusting low-frequency routes. On LLMBar, an
adversarial-heavy construction set transfers cleanly to a natural-heavy batch,
while natural-heavy construction is not the right audit for an
adversarial-heavy next batch. On JBB, proxy-unsafe construction transfers to a
proxy-safe batch and selects a useful routed plan.

\begin{table}[t]
\centering
\scriptsize
\setlength{\tabcolsep}{3.0pt}
\begin{tabular}{lllrrrr}
\toprule
Setting & Construction mix & Deployment mix & Single R/A & Flat R/A & Role R/A & Calls \\
\midrule
LLMBar-7 & natural-heavy & adversarial-heavy & .220/.699 & .237/.646 & .233/.681 & 3.21 \\
LLMBar-7 & adversarial-heavy & natural-heavy & .175/.782 & .215/.686 & .175/.782 & 2.91 \\
JBB-7 & proxy-safe-heavy & proxy-unsafe-heavy & .147/.823 & .154/.803 & .147/.809 & 1.90 \\
JBB-7 & proxy-unsafe-heavy & proxy-safe-heavy & .125/.828 & .147/.724 & .109/.855 & 2.20 \\
\bottomrule
\end{tabular}
\caption{Slice-mix shift audit. Construction and deployment splits are disjoint
and intentionally use different route-signal proportions. Entries are mean
held-out risk/accuracy over 10 random shifted splits. The table is read as a
deployment check on whether the audit labels match the next evaluation batch.}
\label{tab:slice-shift}
\end{table}

Table~\ref{tab:stability-selection} turns the same audit into a deployment
diagnostic. Rather than deploying every split-specific route, the owner can keep calls
that appear in at least half of the calibration splits. LLMBar yields a stable
global pair and a small number of recurring slice specialists. JBB has a stable
global safety judge but no routed proxy specialist above the same frequency
threshold, so the conservative action is to deploy the global call and collect
more proxy-slice audit labels before adding routes.

\begin{table}[t]
\centering
\scriptsize
\setlength{\tabcolsep}{2.0pt}
\begin{tabular}{>{\raggedright\arraybackslash}p{0.15\linewidth}
>{\raggedright\arraybackslash}p{0.39\linewidth}
>{\raggedright\arraybackslash}p{0.39\linewidth}}
\toprule
Setting & Stable calls at 50\% audit & Stable calls at 100\% audit \\
\midrule
LLMBar &
Global: DeepSeek 8/10, Prometheus 7/10. Route: natural Gemma 5/10. &
Global: DeepSeek 10/10, Prometheus 10/10. Routes: GPT-inst Llama 6/10, natural Gemma 6/10, neighbor Llama 5/10. \\
JBB proxy &
Global: Selene 6/10. No route reaches 5/10. &
Global: Selene 8/10. No proxy route reaches 5/10. \\
MBPP public-overfit &
Global: hidden-unit verifier 5/10. &
Global: Prometheus 7/10, DeepSeek 7/10. \\
\bottomrule
\end{tabular}
\caption{Frequency audit for conservative deployment. Calls are listed when they appear in at
least 5/10 random splits. Route entries use \texttt{slice:judge}; unqualified
entries are global calls. Low-frequency routes are not deployment failures;
they are prompts to collect more audit labels before relying on that specialist.
This table is not a statistical guarantee.}
\label{tab:stability-selection}
\end{table}

Table~\ref{tab:production-cost-sweep} replaces uniform call costs with a
production-style cost model and sweeps \(\lambda\). For LLMBar and JBB, the
DeepSeek API anchor has cost 5 and local judges have cost 1. For MBPP, local
LLM judges have cost 3, the DeepSeek API anchor has cost 5, and the hidden-unit
verifier has cost 0.05. Increasing \(\lambda\) asks for a leaner policy under
the same validation-gain objective. The resulting plans keep the same
operational form--global calls, routed specialists, and stopped candidates--but
move along the deployment owner's risk-cost frontier.

\begin{table}[t]
\centering
\scriptsize
\begin{tabular}{lrrrrr}
\toprule
Setting & \(\lambda\) & Risk & Accuracy & Cost & Calls \\
\midrule
LLMBar & 0.000 & 0.1884 & 0.7334 & 7.46 & 3.46 \\
LLMBar & 0.002 & 0.1857 & 0.7468 & 7.08 & 3.08 \\
LLMBar & 0.005 & 0.1871 & 0.7302 & 5.05 & 2.62 \\
JBB proxy & 0.000 & 0.1094 & 0.8527 & 3.09 & 2.29 \\
JBB proxy & 0.002 & 0.1146 & 0.8389 & 1.82 & 1.82 \\
JBB proxy & 0.005 & 0.1169 & 0.8326 & 1.52 & 1.52 \\
MBPP public-overfit & 0.000 & 0.0097 & 0.9900 & 5.91 & 1.70 \\
MBPP public-overfit & 0.002 & 0.0141 & 0.9853 & 3.93 & 1.60 \\
MBPP public-overfit & 0.005 & 0.0224 & 0.9767 & 3.33 & 1.40 \\
\bottomrule
\end{tabular}
\caption{Production-cost sensitivity. Costs are normalized but non-uniform:
API anchors are expensive, local judges are cheaper, and deterministic
verifiers are cheapest. Entries are role-routed policy means over 10 random
splits.}
\label{tab:production-cost-sweep}
\end{table}

\section{Near-Duplicate and Search-Space Audits}
\label{app:near-duplicate-audits}

Exact copies are the easiest redundancy case. A more useful deployment question
is whether prompt variants of the same judge family should be treated as extra
judges. Table~\ref{tab:near-duplicate} forms a ten-call LLMBar pool by pairing
JSON-schema and letter-prompt outputs from five local judge families. The
audit uses the 680 complete-case rows reported in
Table~\ref{tab:complete-case}. The result separates two effects. Calling every
variant is not diversity: flat aggregation over all variants raises risk to
\(0.2633\) and costs ten calls. Yet a prompt variant can still become a slice specialist. The role-routed
policy reaches \(0.7143\) accuracy at \(3.64\) calls, exceeding the full-call
logistic stack accuracy of \(0.6862\) while using fewer calls. The operational
rule is therefore not to delete all near-duplicates, nor to buy all variants;
it is to route variants only where their conditional validation gain clears the
deployment threshold.

\begin{table}[t]
\centering
\scriptsize
\begin{tabular}{lrrr}
\toprule
Policy & Risk \(\pm\) 95\% CI & Accuracy & Cost \\
\midrule
Single best & \(0.2350 \pm 0.0035\) & 0.6126 & 1.00 \\
Flat all prompt variants & \(0.2633 \pm 0.0076\) & 0.5668 & 10.00 \\
Matched top-\(k\) & \(0.2320 \pm 0.0055\) & 0.6335 & 2.10 \\
Frugal confidence cascade & \(0.2351 \pm 0.0086\) & 0.6300 & 1.92 \\
Full-call logistic stack & \(0.2028 \pm 0.0042\) & 0.6862 & 10.00 \\
Role-routed stop & \(0.2122 \pm 0.0080\) & 0.7143 & 3.64 \\
\bottomrule
\end{tabular}
\caption{Near-duplicate prompt-variant audit on LLMBar. The pool contains
JSON-schema and letter-prompt outputs from five judge families. Role routing
keeps prompt variants as conditional specialists rather than treating them as
automatic diversity.}
\label{tab:near-duplicate}
\end{table}

The stopping report can also be audited for pairwise complementarity.
After the greedy path stops, Table~\ref{tab:pair-addition-audit} checks whether
any remaining pair clears the same validation-gain threshold when no remaining
single judge does. LLMBar and MBPP have no pair-only misses across ten splits.
JBB and the prompt-variant audit expose a small number of pair-only moves, so a
deployment owner who wants to search beyond single additions can widen the
stopping report to beam or subset moves on those regimes.

\begin{table}[t]
\centering
\scriptsize
\begin{tabular}{lrrrr}
\toprule
Audit setting & Splits & Pair-only moves & Mean single gain & Max pair gain \\
\midrule
LLMBar, seven judges & 10 & 0 & 0.0008 & 0.0039 \\
MBPP public-overfit & 10 & 0 & -0.0008 & 0.0008 \\
JBB safety & 10 & 3 & 0.0011 & 0.0116 \\
LLMBar prompt variants & 10 & 2 & 0.0011 & 0.0112 \\
\bottomrule
\end{tabular}
\caption{Pair-addition audit after greedy stopping. A pair-only move is a pair
whose validation gain exceeds \(\tau=0.005\) after every remaining single
addition falls below threshold.}
\label{tab:pair-addition-audit}
\end{table}

\section{Stopping Reports}

The method produces auditable stopping decisions. On HumanEval, the first
global step adds the unit-test verifier, reducing construction-validation risk
to zero; all remaining LLM judges then have zero validation gain and the panel
stops. This report is deployment-relevant: if the verifier is available
and matches the target, the default deployment call is the verifier alone.

On LLMBar, the report has a different shape. The global panel adds a small
set of broad complements, then stops globally when the next broad gain falls
below threshold. Slice construction then routes different specialists to
adversarial and natural subsets. This demonstrates that a judge can fail to be
worth invoking globally while still being worth invoking conditionally.

On the redundant-copy stress test, the report states that the injected
copies have no conditional validation gain after the original judge is present.
This is why role risk and role cost are unchanged when the candidate pool grows
from seven to eleven judges.

\end{document}